\documentclass[10pt, a4paper]{article}

\usepackage{lrec-coling2024}

\usepackage{tempora}
\usepackage{listings}

\usepackage{natbib}
\usepackage{multibib}
\makeatletter
\def\@mb@citenamelist{cite,citep,citet,citealp,citealt,citepalias,citetalias}
\makeatother
\newcites{languageresource}{~}

\usepackage{graphicx}
\usepackage{tabularx}
\usepackage{soul}
\usepackage{titlesec}
\titleformat{\section}{\normalfont\large\bfseries\center}{\thesection.}{1em}{}
\titleformat{\subsection}{\normalfont\SmallTitleFont\bfseries\raggedright}{\thesubsection.}{1em}{}
\titleformat{\subsubsection}{\normalfont\normalsize\bfseries\raggedright}{\thesubsubsection.}{1em}{}
\renewcommand\thesection{\arabic{section}}
\renewcommand\thesubsection{\thesection.\arabic{subsection}}
\renewcommand\thesubsubsection{\thesubsection.\arabic{subsubsection}}

\usepackage[hyphens]{url}

\usepackage{breakurl}
\usepackage[breaklinks]{hyperref}

\usepackage[table,xcdraw]{xcolor}
\usepackage{hyperref}
\definecolor{darkblue}{rgb}{0, 0, 0.5}
\hypersetup{colorlinks=true, citecolor=darkblue, linkcolor=darkblue, urlcolor=darkblue}

\usepackage{xstring}

\usepackage{color}

\usepackage{multirow,booktabs}
\usepackage[T2A,T1]{fontenc}
\usepackage[english, russian]{babel}
\usepackage{extdash}
\usepackage{xstring}
\usepackage{tipa}
\usepackage{tabularx,booktabs}
\usepackage{acro}

\usepackage{arydshln}

\usepackage{float}
\restylefloat{table}

\addto\captionsrussian{
\renewcommand{\abstractname}{Abstract}

}

\newcommand{\scell}[2][c]{
\begin{tabular}[#1]{@{}c@{}}#2\end{tabular}}

\newcommand{\scelll}[2][c]{
\begin{tabular}[#1]{@{}l@{}}#2\end{tabular}}

\DeclareAcronym{bleu}{
short = BLEU,
long = bilingual evaluation understudy,
}

\DeclareAcronym{chrf}{
short = chrF,
long = character F-score,
}

\DeclareAcronym{flores}{
short = Facebook Low Resources,
long = FLoRes,
}

\DeclareAcronym{mt}{
short = MT,
long = machine translation,
}

\DeclareAcronym{nlp}{
short = NLP,
long = natural language processing,
}

\DeclareAcronym{nllb}{
short = NLLB,
long = No Language Left Behind,
}

\DeclareAcronym{nmt}{
short = NMT,
long = neural machine translation,
}

\DeclareAcronym{rbmt}{
short = RBMT,
long = rule-based machine translation,
}

\DeclareAcronym{smt}{
short = SMT,
long = statistical machine translation,
}

\DeclareAcronym{ter}{
short = TER,
long = translation edit rate,
}

\DeclareAcronym{wer}{
short = WER,
long = word error rate,
}

\DeclareAcronym{wmt19}{
short = WMT19,
long = Fourth Conference on Machine Translation,
}

\title{KazParC: Kazakh Parallel Corpus for Machine Translation}

\name{Rustem Yeshpanov, Alina Polonskaya, Huseyin Atakan Varol}

\address{Institute of Smart Systems and Artificial Intelligence\\Nazarbayev University, Astana, Kazakhstan \\
\{rustem.yeshpanov, alina.polonskaya, ahvarol\}@nu.edu.kz\\}

\abstract{
We introduce KazParC, a parallel corpus designed for machine translation across Kazakh, English, Russian, and Turkish. The first and largest publicly available corpus of its kind, 
KazParC contains a collection of 371,902 parallel sentences covering different domains and developed with the assistance of human translators. Our research efforts also extend to the development of a neural machine translation model nicknamed Tilmash. Remarkably, the performance of Tilmash is on par with, and in certain instances, surpasses that of industry giants, such as Google Translate and Yandex Translate, as measured by standard evaluation metrics, such as BLEU and chrF. Both KazParC and Tilmash are openly available for download under the Creative Commons Attribution 4.0 International License (CC BY 4.0) through our GitHub repository.
\\\newline
}

\begin{document}

\maketitleabstract

\section{Introduction}
\Ac{mt} refers to the use of computer systems tasked to automatically translate between languages with or without human intervention~\cite{HUTCHINS1995431}. Beyond its fundamental role in linguistic translation, \ac{mt} demonstrates great versatility extending to practical applications in various domains. These applications include accessing and gaining information in another language~\cite{wiesmann2019machine}, language learning and teaching~\cite{lee2020impact}, facilitating professional translation tasks~\cite{craciunescu2004machine}, and providing multilingual customer service~\cite{barrera2016enhancing,lewis2012using}.

\ac{mt} approaches include rule-based, statistical, and neural methods. \Ac{smt} gained ground over \ac{rbmt} in the late 1990s thanks to its ability to learn from large bilingual corpora, making it more adaptable to different language pairs and contexts. However, the dominance of \ac{smt} was challenged by the emergence of \ac{nmt} in the mid-2010s, when \ac{nmt} models with the sequence-to-sequence network~\cite{sutskever2014sequence} displayed unprecedented translation quality and fluency, as well as the ability to handle a wide range of linguistic phenomena~\cite{stahlberg2020neural}, leading to their widespread adoption.

Modern \ac{mt} models are typically trained on large-scale parallel corpora containing pairs of source and target language texts, also known as bitexts~\cite{jurafsky2009}.
Similar to many other domains of \ac{nlp}, \ac{mt} faces a resource imbalance.
While some languages, such as English, Japanese, Mandarin, and Spanish~\cite{koehn-2005-europarl,pryzant2018jesc}, benefit from a wealth of parallel corpora, linguistic tools, and pre-trained models, lower-resourced languages are often in a state of resource paucity, yearning for the abundance available to their higher-resourced counterparts. 

This paper focuses on \ac{nmt} from and to Kazakh, a Turkic language that utilises the Cyrillic script and has an estimated 13 million native speakers~\cite{campbell2020compendium,johanson2021}.
Notwithstanding notable recent advances in Kazakh \ac{nlp}~\cite{mussakhojayeva-etal-2022-kazakhtts2,yeshpanov-etal-2022-kaznerd}, the language remains relatively lower-resourced and in need of further research efforts and resource development, with \ac{mt}, especially in terms of the availability of parallel data, being one of these critical areas.

In this study, we attempt to bridge this source scarcity by presenting a parallel corpus for four languages. The corpus includes parallel data for two Turkic languages, Kazakh and Turkish, belonging to the Kypchak and Oghuz branches, respectively. We also provide parallel data for two Indo-European languages, English and Russian, representing the West-Germanic and Slavic branches, in turn. Furthermore, we introduce an \ac{nmt} model trained using the aforementioned parallel corpus.
The experimental results demonstrate that our model achieves competitive and, in some cases, even superior performance to that of industry giants, when evaluated using standard evaluation metrics, such as \ac{bleu}~\cite{Papineni2002BleuAM} and \ac{chrf}~\cite{popovic-2015-chrf}.

The structure of the paper is as follows: \hyperref[sec2]{Section~2} offers a review of previous research within the field. \hyperref[sec3]{Section~3} delves into the details of data sources, 
collection, pre-processing, partitioning methods, and corpus statistics. \hyperref[sec4]{Section~4} is comprised of subsections focusing on the experimental design, evaluation metrics, and experimental results. \hyperref[sec5]{Section~5} provides a discussion of the obtained results. \hyperref[sec6]{Section~6} concludes the paper and outlines potential areas for future work.

\section{Related Work}\label{sec2}
Kazakhstan implements a trilingual policy, designating Kazakh as its official state language, Russian as the language for interethnic communication, and English as the language essential for effective global economic integration~\cite{sanders2016staying}.
Consequently, most research in Kazakh \ac{mt} has predominantly revolved around Russian or English as either source or target translation languages.

Early attempts at Kazakh$\leftrightarrow$English and Kazakh$\leftrightarrow$Russian \ac{mt} involved building structural transfer rules on Apertium~\cite{forcada2011apertium,6522002,sundetova2014structural}, implementing morphological segmentation techniques to address the rich morphology of the Kazakh language~\cite{assylbekov2014initial,bekbulatov2014study}, and exploring sentence alignment through Russian lemmatisation and bilingual dictionaries~\cite{assylbekov2016experiments,myrzakhmetov2016initial}.

Regarding Kazakh$\leftrightarrow$Turkish \ac{mt}, the scarcity of parallel training data has posed a significant limitation, resulting in a small number of research studies dedicated to the development of translation systems for these two Turkic languages~\cite{10.1145/3443279.3443286}. As an illustration, in the study by~\citet{Bayatli2018RulebasedMT}, efforts were made to address this data deficit by manually translating about a thousand Kazakh treebank sentences~\cite{tyers2015towards} into Turkish to create an \ac{rbmt} system. This system achieved a \ac{bleu} score of 0.17 and \ac{wer} of 0.46.

It is worth noting that parallel data for the aforementioned language pairs did exist to some extent~\cite{TIEDEMANN12.463}. However, the prevailing approach in most related studies was to create custom parallel corpora~\cite{kuandykova2014english}. This practice was motivated by the numerous issues in the existing data, including recurring repetitions, corrupted text segments, and obvious misalignment between the pairs~\cite{myrzakhmetov2016initial}, which collectively contributed to a substantial reduction in the quality and quantity of the available data.

In~\citet{Rakhimova2017}, Kazakh--English (25,000 sentences) and Kazakh--Russian (10,000 sentences) parallel corpora were constructed utilising an open-source tool designed for the extraction of bitexts from multilingual websites. In a separate study by~\citet{10.1007/978-3-319-67077-5_48}, the researchers collected an additional 73,031 Kazakh--English parallel sentences using the same tool. Importantly, the data collected in both studies were aligned automatically and are not open access.

In~\citet{makazhanov}, over 890,000 parallel sentences in Russian and Kazakh were extracted from online news articles published on websites related to state institutions, national companies, and other quasi-governmental bodies. An \ac{smt} model trained on the parallel data yielded a \ac{bleu} score of 0.34. Interestingly, in~\citet{doi:10.1080/23311916.2020.1856500}, the authors achieved \ac{bleu} scores of 0.25 and 0.18 for the Kazakh$\rightarrow$English and English$\rightarrow$Kazakh language pairs, respectively, training an \ac{nmt} model on a dataset acquired from the same aforementioned sources, although more than eight times smaller in size.
In their later study~\cite{refId0}, a 439,176-sentence-long synthetic corpus using the complete set of Kazakh suffixes was constructed. An \ac{nmt} model trained on the corpus produced \ac{bleu} scores in the range of 0.14 to 0.16 for the Kazakh$\leftrightarrow$Russian and Kazakh$\leftrightarrow$English language pairs.

In the study by~\citet{khairova2019aligned}, automated alignment was performed to create a Kazakh-Russian parallel corpus. This corpus comprised 3,000 texts that were extracted from four bilingual news websites in Kazakhstan, with a specific focus on criminal-related content. The researchers acknowledged the intricate syntactic structures inherent in both Kazakh and Russian, which posed significant challenges to the automatic alignment process. It was further observed that approximately 40\% of the sentences required manual alignment due to these complexities.

\begin{table*}[h]
\centering
\setlength\tabcolsep{0.13cm}

\begin{tabularx}\textwidth{crrrrrrrrrr}
\toprule
\multirow{3}{*}{\textbf{Domain}} & \multicolumn{2}{c}{\multirow{2}{*}{\textbf{lines}}}
& \multicolumn{8}{c}{\textbf{tokens}} \\
&  & 
& \multicolumn{2}{c}{\textbf{EN}}
& \multicolumn{2}{c}{\textbf{KK}}
& \multicolumn{2}{c}{\textbf{RU}}
& \multicolumn{2}{c}{\textbf{TR}}
\\
& \multicolumn{1}{c}{\textbf{\#}} & \multicolumn{1}{c}{\textbf{\%}} &\multicolumn{1}{c}{\textbf{\#}} & 
\multicolumn{1}{c}{\textbf{\%}}
& \multicolumn{1}{c}{\textbf{\#}} & 
\multicolumn{1}{c}{\textbf{\%}}
& \multicolumn{1}{c}{\textbf{\#}} & 
\multicolumn{1}{c}{\textbf{\%}}
& \multicolumn{1}{c}{\textbf{\#}} & 
\multicolumn{1}{c}{\textbf{\%}}
\\
\midrule
Mass media & 120,547 & 32.4 & 1,817,276 & 28.3 & 1,340,346 & 28.6 & 1,454,430 & 29.0 & 1,311,985 & 28.5 \\
General & 94,988 & 25.5 & 844,541 & 13.1 & 578,236 & 12.3 & 618,960 & 12.3 & 608,020 & 13.2 \\
Legal documents & 77,183 & 20.8 & 2,650,626 & 41.3 & 1,925,561 & 41.0 & 1,991,222 & 39.7 & 1,880,081 & 40.8\\
Education and science & 46,252 & 12.4 & 522,830 & 8.1 & 392,348 & 8.4 & 444,786 & 8.9 & 376,484 & 8.2\\
Fiction & 32,932 & 8.9 & 589,001 & 9.2 & 456,385 & 9.7 & 510,168 & 10.2 & 433,968 & 9.4\\   
\midrule
\textbf{Total} & \textbf{371,902} & \textbf{100} & \textbf{6,424,274} & \textbf{100} & \textbf{4,692,876} & \textbf{100} & \textbf{5,019,566} & \textbf{100} & \textbf{4,610,538} & \textbf{100}\\
\bottomrule
\end{tabularx}
\caption{KazParC domain statistics}
\label{tab:kazparc_domain_stats}
\end{table*}

The inclusion of the Kazakh$\leftrightarrow$English language pair as a translation task within the \ac{wmt19} sparked several research efforts.
Given the limited availability of parallel data for Kazakh--English, these initiatives leveraged the more abundant English--Russian and Kazakh--Russian sentence pairs, which numbered approximately 14 million and 5 million, respectively, using Russian as a pivot language~\cite{casas-etal-2019-talp,littell-etal-2019-multi,sanchez-cartagena-etal-2019-universitat}. Additional attempts involved transfer learning utilising supplementary parallel data from the Turkish$\leftrightarrow$English language pair, as Turkish shares linguistic kinship with Kazakh~\cite{toral-etal-2019-neural,Briakou2019TheUO}, albeit being a low-resource language itself.
While~\citet{Briakou2019TheUO} obtained a \ac{bleu} score of 0.1 with just over 100 thousand Kazakh--English sentence pairs and another 200 thousand sentence pairs from Turkish--English data,~\citet{toral-etal-2019-neural}, using English--Russian, Kazakh--Russian, and English--Turkish data, achieved a \ac{bleu} of 0.24 for Kazakh$\rightarrow$English and a \ac{chrf} of 0.48 for English$\rightarrow$Kazakh.

While recent research efforts in Kazakh$\leftrightarrow$English and Kazakh$\leftrightarrow$Russian \ac{mt} have demonstrated noteworthy advancements, including the development of large-scale crawled parallel corpora~\cite{rakhimova2022aligning,10.1007/978-3-030-88113-9_41}, which are publicly accessible and capable of yielding commendable \ac{bleu} scores of up to 0.49~\cite{karyukin2023neural}, as well as the construction of \ac{nmt} post-editing models trained on such data~\cite{10.1007/978-3-030-88113-9_42}, the majority of textual resources continue to come from governmental sources. 
This preference is attributed to the perception of governmental texts as subjected to moderation and therefore trustworthy~\cite{karyukin2023neural}. 
However, it should be noted that ensuring the quality and alignment of such texts still requires a significant amount of manual intervention~\cite{10.1007/978-3-030-88113-9_41}. 
Excessive reliance on sources related to state bodies further harbours the potential to introduce bias into the corpus, thereby constraining the generalisability of models trained on such data. 
In light of these challenges, our study sought to create an extensive parallel corpus containing texts from diverse sources through the collaborative contributions of human translators, which would hopefully facilitate \ac{mt} across Kazakh, English, Russian, and Turkish, as elaborated in subsequent sections.

\section{Corpus Development}\label{sec3}
\subsection{Data Sources}\label{sec3.1}
The data for our \textbf{Kaz}akh \textbf{Par}allel \textbf{C}orpus (hereafter KazParC) were sourced from a wide selection of textual materials, including proverbs and sayings, terminology glossaries, phrasebooks, literary works, periodicals, language learning resources, including the SCoRE corpus~\citep{chujo2015corpus}, educational video subtitle collections, such as QED~\citep{abdelali-etal-2014-amara}, news items, such as KazNERD~\cite{yeshpanov-etal-2022-kaznerd} and WMT~\citep{TIEDEMANN12.463}, TED talks\footnote{\url{https://www.ted.com/}}, governmental and regulatory legal documents from Kazakhstan\footnote{\url{https://adilet.zan.kz/}}, communications from the official website of the President of the Republic of Kazakhstan\footnote{\url{https://www.akorda.kz/}}, United Nations publications\footnote{\url{https://www.un.org/}}, and image captions derived from sources, such as COCO~\citep{10.1007/978-3-319-10602-1_48}. The data acquired from these sources were subsumed under five broad categories or domains---namely, Education and science, Fiction, General, Legal documents, and Mass media. 
Table~\ref{tab:kazparc_domain_stats} provides information about the number of lines and tokens collected per domain.

\subsection{Data Collection}
The process of data collection, which involved gathering text materials and their translation, was initiated in July 2021 and persisted until September 2023.
Throughout this period, an average of 10 human translators were involved, which equates to 41,600 hours of human effort (26 months x 10 translators x 160 hours/month). The human translators not only engaged in the collection of readily translated publicly available data but also undertook the translation of texts that originally lacked translations in the languages under consideration.

The data collected were screened to remove any information that could potentially identify individuals, as well as to filter out instances of hate speech, discriminatory language, or violence.
Subsequently, the data were segmented into sentences, each labelled with a domain identifier. A careful review for grammatical and spelling accuracy was conducted and duplicate sentences removed. Given the common practice of Kazakh-Russian code-switching in Kazakhstan~\citep{Pavlenko2008}, sentences containing both Kazakh and Russian words underwent a modification process, wherein the Russian elements were translated into Kazakh for uniformity, taking care not to compromise the intended meaning of the sentences.

\subsection{Data Pre-Processing}
All the data collected were subjected to initial pre-processing, which involved segmenting the data into language pairs. 
Extraneous characters were systematically eliminated and homoglyphs effectively replaced. 
In addition, the characters responsible for line breaks (\textbackslash n) and carriage returns (\textbackslash r) were removed. 
The pre-processing further entailed the identification and elimination of duplicate entries, filtering out rows with identical text in both language columns. 
However, in order to enrich the diversity of the corpus and capture a wider range of synonyms for different words and expressions, lines with duplicate text in a single language column were judiciously retained.

In Table~\ref{tab:kazparc_pairwise_stats}, we present statistics for language pairs within the corpus. 
The ``\# lines'' column indicates the number of rows per language pair. In the `\# sents'', ``\# tokens'', ``\# types'' columns, we provide unique sentence, token, and type (i.e., unique token) counts for each language pair, respectively, with the upper numbers referring to the first language in the pair and the lower numbers to the second language.
The token and type counts were obtained after processing the data with Moses tokeniser 1.2.1\footnote{\url{https://pypi.org/project/mosestokenizer/}}.

\begin{table}[!t]
\fontsize{10}{12}\selectfont
\setlength\tabcolsep{0.13cm}
\begin{tabularx}\columnwidth{ccccc}
\toprule
\textbf{Pair} & \textbf{\scell{\#\\lines}} & \textbf{\scell{\#\\sents}} & \textbf{\scell{\#\\tokens}} & \textbf{\scell{\#\\types}} \\
\midrule
KK$\leftrightarrow$EN & 363,594 & \scell{362,230\\361,087} & \scell{4,670,789\\6,393,381} & \scell{184,258\\59,062} \\
\hline
KK$\leftrightarrow$RU & 363,482 & \scell{362,230\\362,748} & \scell{4,670,593\\4,996,031} & \scell{184,258\\183,204}  \\
\hline
KK$\leftrightarrow$TR & 362,150 & \scell{362,230\\361,660} & \scell{4,668,852\\4,586,421} & \scell{184,258\\175,145}  \\
\hline
EN$\leftrightarrow$RU & 363,456 & \scell{361,087\\362,748} & \scell{6,392,301\\4,994,310} & \scell{59,062\\183,204}  \\
\hline
EN$\leftrightarrow$TR & 362,392 & \scell{361,087\\361,660} & \scell{6,380,703\\4,579,375} & \scell{59,062\\175,145}  \\
\hline
RU$\leftrightarrow$TR & 363,324 & \scell{362,748\\361,660} & \scell{4,999,850\\4,591,847} & \scell{183,204\\175,145}  \\
\bottomrule
\end{tabularx}
\caption{KazParC pairwise statistics}
\label{tab:kazparc_pairwise_stats}
\end{table}

\subsection{Data Splitting}
We first created a test set. 
To this end, we conducted a random selection process, curating a set comprising 250 distinct and non-repetitive rows from each of the specified sources in \hyperref[sec3.1]{Section~3.1}. 
The remaining data were partitioned pairwise in adherence to an 80/20 ratio, preserving the distribution of domains within the training and validation sets (see Table~\ref{tab:kazparc_set_stats}). 

\subsection{Synthetic Corpus}
To expand the scope of our parallel corpus and enhance its data diversity, as well as to investigate the performance characteristics of the developed \ac{nmt} models when confronted with a combination of human-translated and machine-translated content, we conducted web crawling to acquire a total of 1,797,066 sentences from English-language websites. Subsequently, these sentences underwent an automated translation process into Kazakh, Russian, and Turkish languages utilising the Google Translate service. Within the context of our research, this collection of data will be referred to as ``SynC'' (\textbf{Syn}thetic \textbf{C}orpus). Table~\ref{tab:synthetic_pairwise_stats} presents statistics pertaining to the quantity of unique sentences, tokens, and types per each language pair.
The synthetic corpus was further partitioned pairwise into training and validation sets at a ratio of 90/10 to facilitate model development and evaluation (see Table~\ref{tab:synthetic_sets}).

\begin{table}[!t]
\fontsize{9.3}{12}\selectfont
\setlength\tabcolsep{0.11cm}
\begin{tabularx}\columnwidth{ccccc}
\toprule
\textbf{Pair} & \textbf{\scell{\#\\lines}} & \textbf{\scell{\#\\sents}} & \textbf{\scell{\#\\tokens}} & \textbf{\scell{\#\\types}} \\
\midrule
KK$\leftrightarrow$EN & 1,787,050 & \scell{1,782,192\\1,781,019} & \scell{26,630,960\\35,291,705} & \scell{685,135\\300,556} \\
\hline
KK$\leftrightarrow$RU & 1,787,448 & \scell{1,782,192\\1,777,500} & \scell{26,654,195\\30,241,895} & \scell{685,135\\672,146}  \\
\hline
KK$\leftrightarrow$TR & 1,791,425 & \scell{1,782,192\\1,782,257} & \scell{26,726,439\\27,865,860} & \scell{685,135\\656,294}  \\
\hline
EN$\leftrightarrow$RU & 1,784,513 & \scell{1,781,019\\1,777,500} & \scell{35,244,800\\30,175,611} & \scell{300,556\\672,146}  \\
\hline
EN$\leftrightarrow$TR & 1,788,564 & \scell{1,781,019\\1,782,257} & \scell{35,344,188\\27,806,708} & \scell{300,556\\656,294}  \\
\hline
RU$\leftrightarrow$TR & 1,788,027 & \scell{1,777,500\\1,782,257} & \scell{30,269,083\\27,816,210} & \scell{672,146\\656,294}  \\
\bottomrule
\end{tabularx}
\caption{SynC pairwise statistics}
\label{tab:synthetic_pairwise_stats}
\end{table}

\begin{table*}[!t]
\fontsize{9}{12}\selectfont
\setlength\tabcolsep{0.085cm}
\begin{tabularx}\textwidth{ccccccccccccccc}
\toprule
\multirow{2}{*}{\textbf{Pair}} & \multicolumn{4}{c}{\textbf{Train}} &  & \multicolumn{4}{c}{\textbf{Valid}} &  & \multicolumn{4}{c}{\textbf{Test}} \\
\cmidrule{2-5} \cmidrule{7-10} \cmidrule{12-15}
& \textbf{\# lines} & \textbf{\# sents} & \textbf{\# tokens} & \textbf{\# types} &  & \textbf{\# lines} & \textbf{\# sents} & \textbf{\# tokens} & \textbf{\# types} &  & \textbf{\# lines} & \textbf{\# sents} &\textbf{ \# tokens} & \textbf{\# types} \\
\midrule
KK$\leftrightarrow$EN & 290,877 & \begin{tabular}[c]{@{}c@{}}286,958\\ 286,197\end{tabular} & \begin{tabular}[c]{@{}c@{}}3,693,263\\ 5,057,687\end{tabular} & \begin{tabular}[c]{@{}c@{}}164,766\\ 54,311\end{tabular} &  & 72,719 & \begin{tabular}[c]{@{}c@{}}72,426 \\ 72,403\end{tabular} & \begin{tabular}[c]{@{}c@{}}920,482\\ 1,259,827\end{tabular} & \begin{tabular}[c]{@{}c@{}}83,057\\ 32,063\end{tabular} &  & 4,750 & \begin{tabular}[c]{@{}c@{}}4,750 \\ 4,750\end{tabular} & \begin{tabular}[c]{@{}c@{}}57,044\\ 75,867\end{tabular} & \begin{tabular}[c]{@{}c@{}}17,475\\ 9,729\end{tabular} \\
\hline
KK$\leftrightarrow$RU & 290,785 & \begin{tabular}[c]{@{}c@{}}286,943 \\ 287,215\end{tabular} & \begin{tabular}[c]{@{}c@{}}3,689,799\\ 3,945,741\end{tabular} & \begin{tabular}[c]{@{}c@{}}164,995\\ 165,882\end{tabular} &  & 72,697 & \begin{tabular}[c]{@{}c@{}}72,413\\ 72,439\end{tabular} & \begin{tabular}[c]{@{}c@{}}923,750\\ 988,374\end{tabular} & \begin{tabular}[c]{@{}c@{}}82,958\\ 87,519\end{tabular} &  & 4,750 & \begin{tabular}[c]{@{}c@{}}4,750 \\ 4,750\end{tabular} & \begin{tabular}[c]{@{}c@{}}57,044\\ 61,916\end{tabular} & \begin{tabular}[c]{@{}c@{}}17,475\\ 18,804\end{tabular} \\
\hline
KK$\leftrightarrow$TR & 289,720 & \begin{tabular}[c]{@{}c@{}}286,694 \\ 286,279\end{tabular} & \begin{tabular}[c]{@{}c@{}}3,691,751\\ 3,626,361\end{tabular} & \begin{tabular}[c]{@{}c@{}}164,961\\ 157,460\end{tabular} &  & 72,430 & \begin{tabular}[c]{@{}c@{}}72,211 \\ 72,190\end{tabular} & \begin{tabular}[c]{@{}c@{}}920,057\\ 904,199\end{tabular} & \begin{tabular}[c]{@{}c@{}}82,698\\ 80,885\end{tabular} &  & 4,750 & \begin{tabular}[c]{@{}c@{}}4,750 \\ 4,750\end{tabular} & \begin{tabular}[c]{@{}c@{}}57,044\\ 55,861\end{tabular} & \begin{tabular}[c]{@{}c@{}}17,475\\ 17,284\end{tabular} \\
\hline
EN$\leftrightarrow$RU & 290,764 & \begin{tabular}[c]{@{}c@{}}286,185 \\ 287,261\end{tabular} & \begin{tabular}[c]{@{}c@{}}5,058,530\\ 3,950,362\end{tabular} & \begin{tabular}[c]{@{}c@{}}54,322\\ 165,701\end{tabular} &  & 72,692 & \begin{tabular}[c]{@{}c@{}}72,377 \\ 72,427\end{tabular} & \begin{tabular}[c]{@{}c@{}}1,257,904\\ 982,032\end{tabular} & \begin{tabular}[c]{@{}c@{}}32,208\\ 87,541\end{tabular} &  & 4,750 & \begin{tabular}[c]{@{}c@{}}4,750 \\ 4,750\end{tabular} & \begin{tabular}[c]{@{}c@{}}75,867\\ 61,916\end{tabular} & \begin{tabular}[c]{@{}c@{}}9,729\\ 18,804\end{tabular} \\
\hline
EN$\leftrightarrow$TR & 289,913 & \begin{tabular}[c]{@{}c@{}}285,967\\ 286,288\end{tabular} & \begin{tabular}[c]{@{}c@{}}5,048,274\\ 3,621,531\end{tabular} & \begin{tabular}[c]{@{}c@{}}54,224\\ 157,369\end{tabular} &  & 72,479 & \begin{tabular}[c]{@{}c@{}}72,220 \\ 72,219\end{tabular} & \begin{tabular}[c]{@{}c@{}}1,256,562\\ 901,983\end{tabular} & \begin{tabular}[c]{@{}c@{}}32,269\\ 80,838\end{tabular} &  & 4,750 & \begin{tabular}[c]{@{}c@{}}4,750 \\ 4,750\end{tabular} & \begin{tabular}[c]{@{}c@{}}75,867\\ 55,861\end{tabular} & \begin{tabular}[c]{@{}c@{}}9,729\\ 17,284\end{tabular} \\
\hline
RU$\leftrightarrow$TR & 290,899 & \begin{tabular}[c]{@{}c@{}}287,241 \\ 286,475\end{tabular} & \begin{tabular}[c]{@{}c@{}}3,947,809\\ 3,626,436\end{tabular} & \begin{tabular}[c]{@{}c@{}}165,482\\ 157,470\end{tabular} &  & 72,725 & \begin{tabular}[c]{@{}c@{}}72,455\\ 72,362\end{tabular} & \begin{tabular}[c]{@{}c@{}}990,125\\ 909,550\end{tabular} & \begin{tabular}[c]{@{}c@{}}87,831\\ 80,962\end{tabular} &  & 4,750 & \begin{tabular}[c]{@{}c@{}}4,750 \\ 4,750\end{tabular} & \begin{tabular}[c]{@{}c@{}}61,916\\ 55,861\end{tabular} & \begin{tabular}[c]{@{}c@{}}18,804\\ 17,284\end{tabular}\\
\bottomrule
\end{tabularx}
\caption{KazParC training, validation, and test sets (by line, sentence, token, and type)}
\label{tab:kazparc_set_stats}
\end{table*}

\begin{table*}[!h]
\fontsize{10}{12}\selectfont
\setlength\tabcolsep{0.215cm}
\begin{tabularx}\textwidth{cccccccccc}
\toprule
\multirow{2}{*}{\textbf{Pair}} & \multicolumn{4}{c}{\textbf{Train}} &  & \multicolumn{4}{c}{\textbf{Valid}} \\
\cmidrule{2-5} \cmidrule{7-10}
& \textbf{\# lines} & \textbf{\# sents} & \textbf{\# tokens} & \textbf{\# types} &  & \textbf{\# lines} & \textbf{\# sents} & \textbf{\# tokens} & \textbf{\# types} \\
\midrule
\multirow{2}{*}{KK$\leftrightarrow$EN} & \multirow{2}{*}{1,608,345} & 1,604,414 & 23,970,260 & 650,144 &  & \multirow{2}{*}{178,705} & 178,654 & 2,660,700 & 208,838 \\
&  & 1,603,426 & 31,767,617 & 286,372 &  &  & 178,639 & 3,524,088 & 105,517 \\
\hline
\multirow{2}{*}{KK$\leftrightarrow$RU} & \multirow{2}{*}{1,608,703} & 1,604,468 & 23,992,148 & 650,170 &  & \multirow{2}{*}{178,745} & 178,691 & 2,662,047 & 209,188 \\
&  & 1,600,643 & 27,221,583 & 642,604 &  &  & 178,642 & 3,020,312 & 235,642 \\
\hline
\multirow{2}{*}{KK$\leftrightarrow$TR} & \multirow{2}{*}{1,612,282} & 1,604,793 & 24,053,671 & 650,384 &  & \multirow{2}{*}{179,143} & 179,057 & 2,672,768 & 209,549 \\
&  & 1,604,822 & 25,078,688 & 626,724 &  &  & 179,057 & 2,787,172 & 221,773 \\
\hline
\multirow{2}{*}{EN$\leftrightarrow$RU} & \multirow{2}{*}{1,606,061} & 1,603,199 & 31,719,781 & 286,645 &  & \multirow{2}{*}{178,452} & 178,419 & 3,525,019 & 104,834 \\
&  & 1,600,372 & 27,158,101 & 642,686 &  &  & 178,379 & 3,017,510 & 235,069 \\
\hline
\multirow{2}{*}{EN$\leftrightarrow$TR} & \multirow{2}{*}{1,609,707} & 1,603,636 & 31,805,393 & 286,387 &  & \multirow{2}{*}{178,857} & 178,775 & 3,538,795 & 105,641 \\
&  & 1,604,545 & 25,022,782 & 626,740 &  &  & 178,796 & 2,783,926 & 221,372 \\
\hline
\multirow{2}{*}{RU$\leftrightarrow$TR} & \multirow{2}{*}{1,609,224} & 1,600,605 & 27,243,278 & 642,797 &  & \multirow{2}{*}{178,803} & 178,695 & 3,025,805 & 235,970 \\
&  & 1,604,521 & 25,035,274 & 626,587 &  &  & 178,750 & 2,780,936 & 221,792 \\
\bottomrule
\end{tabularx}
\caption{SynC: training and validation sets (by line, sentence, token, and type)}
\label{tab:synthetic_sets}
\end{table*}

\subsection{Corpus Structure}
Both KazParC and SynC are openly accessible to the research community through our GitHub repository.\footnote{\url{https://github.com/IS2AI/KazParC}\label{ft:github}}
The corpora consist of multiple files categorised into two distinct groups based on their file prefixes: Files ``01'' through ``19'' bear the ``kazparc'' prefix, while Files ``20'' to ``32'' are denoted by the ``sync'' prefix.

File ``01'' contains the original, unprocessed text collected for the four languages considered within KazParC. Files ``02'' through ``19'' represent pre-processed texts divided into language pairs to serve as training data (Files ``02'' to ``07''), validation data (Files ``08'' to ``13''), and testing data (Files ``14'' to ``19''). Language pairs are denoted within the filenames through the utilisation of two-letter language codes (e.g., kk\_en).

SynC files are organised similarly. File ``20'' holds raw, unprocessed text data from the four languages. Files ``21'' to ``32'' contain pre-processed text split language pairwise for training (Files ``21'' to ``26'') and validation (Files ``27'' to ``32'') purposes.

In Files ``01'' and ``20'', each line comprises distinct components: a unique line identifier (\texttt{id}), texts in Kazakh (\texttt{kk}), English (\texttt{en}), Russian (\texttt{ru}), and Turkish (\texttt{tr}), along with accompanying domain information (\texttt{domain}). As for the remaining files, the data fields are \texttt{id}, \texttt{source\_lang}, \texttt{target\_lang}, \texttt{domain}, and the language \texttt{pair} (e.g., kk\_en).

\vspace{-0.1cm}
\section{Experiment}\label{sec4}

\subsection{Experimental Setup}
The Transformer architecture~\cite{NIPS2017_3f5ee243} has proven highly effective in various \ac{nlp} tasks, including \ac{mt}, text generation, and text classification.
In our study, we opted to employ Facebook's \ac{nllb} model~\citep{nllbteam2022language}. The model supports \ac{mt} for 202 languages, including Kazakh, English, Russian, and Turkish.

We first tested both the baseline\footnote{\url{https://huggingface.co/facebook/nllb-200-1.3B}} and distilled\footnote{\url{https://huggingface.co/facebook/nllb-200-distilled-1.3B}} versions of the model, obtained from the Hugging Face~\citep{wolf-etal-2020-transformers} repository, by fine-tuning them on KazParC. 
Upon comparison of the results, we observed that the distilled model consistently outperformed the baseline model, albeit by a slight margin of 0.01 \ac{bleu}. Therefore, in the subsequent experiments, we focused exclusively on fine-tuning the distilled model.

A total of four models, with each serving a specific purpose, were explored: (1) \texttt{base}, the off-the-shelf model, (2) \texttt{parc}, fine-tuned on KazParC data, (3) \texttt{sync}, fine-tuned on SynC data, and (4) \texttt{parsync}, fine-tuned on both KazParC and SynC data.

The \texttt{base} model was used as a reference point for evaluating the performance of the NLLB model.
The \texttt{parc} model was fine-tuned exclusively on clean, manually translated data and was therefore considered suitable for tasks where accurate translation is important, especially in the domains covered by the training set.

The decision to test a model fine-tuned solely on synthetic data pursued the aim of discerning whether the performance of the model is more influenced by the quality or quantity of data within the parallel corpus. As a result, the \texttt{sync} model was expected to emphasise the viability of using synthetic data in scenarios where creating a human-translated parallel corpus is not feasible.

To assess the influence of data volume on translation quality, we explored the incorporation of synthetic data into our training set.
This investigation aimed not only to evaluate its potential for enhancing translation quality but also to introduce distinctive lexemes absent in the original KazParC.
Therefore, the \texttt{parsync} model was anticipated to leverage the synthetic and manual corpora and achieve a higher degree of universality and applicability to real-world problems.

The hyperparameters were tuned using the validation sets. 
Synthetic data were included in the validation sets only when the performance of the \texttt{sync} and \texttt{parsync} models was assessed. 
The best-performing models were evaluated on the test sets.
Furthermore, we utilised Google Translate\footnote{\url{https://translate.google.com/}} and Yandex Translate\footnote{\url{https://translate.yandex.com/}} to translate the test sets, allowing us to make a comparative assessment between the results generated by our models and those produced by industry-leading machine translation services.
In addition to the KazParС test set, we used the parallel FLoRes-200 (hereafter FLoRes) dataset~\citep{nllbteam2022language}. This dataset was created to evaluate translation quality for 204 languages and contains texts from the Wikivoyage, Wikijunior, and Wikinews resources. FLoRes is divided into \textit{dev} and \textit{devtest} sets, but we combined them into one set.
We also used the FLoRes test set to evaluate the quality for the language pairs German-French (two Latin-based higher-resourced Indo-European languages), German-Ukrainian (a higher-resourced language and a Cyrillic-based lower-resourced Indo-European language), and French-Uzbek (a higher-resourced language and a Latin-based low-resourced Turkic language) to see whether the translation quality changes for these control pairs after fine-tuning the model.

All the models were fine-tuned using eight GPUs on an NVIDIA DGX A100 machine. An initial learning rate of $2\cdot10^{-5}$ was set. The optimization algorithm chosen was AdaFactor. The training spanned across three epochs, with both the training and evaluation batch sizes set to 8.

\subsection{Evaluation Metrics}
In evaluating the \ac{mt} models, we employed two widely recognised metrics: \ac{bleu}~\cite{Papineni2002BleuAM} and \ac{chrf}~\cite{popovic-2015-chrf}. While \ac{bleu} quantifies how closely the machine-produced translation matches human references, by calculating precision in n-grams (4 in our study), \ac{chrf} evaluates translation quality by considering character n-grams instead of word-based approaches. 
This makes \ac{chrf} particularly suitable for agglutinative languages, such as Kazakh and Turkish, which have rich and complex inflectional and derivational morphologies \citep{stanojevic-etal-2015-results}. \ac{chrf} computes the harmonic mean of character-based precision and recall, providing a robust evaluation of translation performance.
Both \ac{bleu} and \ac{chrf} provide a score between 0 and 1, with higher scores indicating better translation quality.

\subsection{Experiment Results}
Model performance results are presented in Table~\ref{tab:bleu_results}.
The table illustrates a notable disparity in bidirectional translation outcomes, particularly between higher-resourced Indo-European languages---English and Russian---and Turkic languages, Kazakh and Turkish. As can be seen from the table, it is apparent that \ac{bleu} scores exhibit a strong and positive correlation with \ac{chrf} scores.

In the ``$\rightarrow$English'' translation direction, Google consistently led on the FLoRes test set, achieving a minimum \ac{bleu} score of 0.35. However, on the KazParC test set, the leadership shifted to the \texttt{parc} model, which was exclusively trained on our parallel corpus. Notably, \texttt{parc} demonstrated an impressive \ac{bleu} score of up to 0.43 when translating RU$\rightarrow$EN.

In the ``$\rightarrow$RU'' translation, Google achieved the highest \ac{bleu} scores on both test sets. The only exception was observed in the EN$\rightarrow$RU translation on the FLoRes test set, where Yandex outperformed Google by a margin of 0.01. Interestingly, when translating ``$\rightarrow$RU'', the \texttt{parc} model generally exhibited lower performance compared to the \texttt{parsync} model, which was trained on a combination of our parallel corpus and synthetic data.

The same pattern was observed for the ``$\rightarrow$KK'' and ``$\rightarrow$TR'' translations. Google obtained the highest \ac{bleu} scores in both test sets. What is truly noteworthy is the clear underperformance of \texttt{parc} compared to \texttt{parsync} in these translation directions. This observation strongly supports the idea that model performance for lower-resourced (Turkic languages) can be substantially enhanced when synthetic data are employed alongside human-translated parallel data.

In the ``EN$\rightarrow$'' translation direction, Google delivered superior translations across both test sets, with exceptions observed where Yandex briefly outperformed in the EN$\rightarrow$RU language pair within the FLoRes dataset. It is worth noting that the \texttt{parsync} model consistently ranked among the top three performers on both test sets, attaining a commendable \ac{bleu} score of 0.20 in the EN$\rightarrow$KK language pair within the FLoRes dataset, a result akin to that of Google.

\begin{table*}[ht]
\fontsize{8}{10}\selectfont
\setlength\tabcolsep{0.075cm}
\begin{tabularx}\textwidth{cccccccccccccc}
\toprule
\multirow{2}{*}{\textbf{Pair}} & \multicolumn{6}{c}{\textbf{FLoRes Test Set}} & & \multicolumn{6}{c}{\textbf{KazParC Test Set}} \\
\cmidrule{2-7} \cmidrule{9-14}
& \texttt{base} & \texttt{parc} & \texttt{sync} & \texttt{\scell{parsync}} & Yandex & Google & & \texttt{base} & \texttt{parc} & \texttt{sync} & \texttt{\scell{parsync}} & Yandex & Google \\
\midrule
EN$\rightarrow$KK & 0.11|0.49 & 0.14|0.56 & \textbf{0.20}|\textbf{0.60} & \textbf{0.20}|\textbf{0.60} & 0.18|0.58 & \textbf{0.20}|\textbf{0.60} & & 0.12|0.51 & 0.18|0.58 & 0.18|0.58 & 0.21|0.60 & 0.18|0.58 & \textbf{0.30}|\textbf{0.65} \\
EN$\rightarrow$RU & 0.25|0.56 & 0.26|0.58 & 0.28|0.60 & 0.28|0.60 & \textbf{0.32}|\textbf{0.63} & 0.31|0.62 & & 0.31|0.64 & 0.38|0.68 & 0.35|0.66 & 0.38|0.68 & 0.39|0.70 & \textbf{0.41}|\textbf{0.71} \\ 
EN$\rightarrow$TR & 0.19|0.58 & 0.22|0.61 & 0.27|0.65 & 0.27|0.65 & 0.29|\textbf{0.66} & \textbf{0.30}|\textbf{0.66} & & 0.19|0.59 & 0.22|0.62 & 0.25|0.63 & 0.25|0.64 & 0.27|0.64 & \textbf{0.34}|\textbf{0.68} \\
KK$\rightarrow$EN & 0.28|0.59 & 0.32|0.62 & 0.31|0.62 & 0.32|0.63 & 0.30|0.62 & \textbf{0.36}|\textbf{0.65} & & 0.24|0.55 & \textbf{0.33}|\textbf{0.62} & 0.24|0.57 & 0.32|\textbf{0.62} & 0.28|0.60 & 0.31|\textbf{0.62} \\
KK$\rightarrow$RU & 0.15|0.49 & 0.17|0.51 & 0.18|0.52 & 0.18|0.52 & 0.18|0.52 & \textbf{0.20}|\textbf{0.53} & & 0.22|0.56 & \textbf{0.29}|\textbf{0.63} & 0.24|0.59 & \textbf{0.29}|\textbf{0.63} & \textbf{0.29}|\textbf{0.63} & \textbf{0.29}|0.61 \\ 
KK$\rightarrow$TR & 0.09|0.48 & 0.13|0.52 & 0.14|0.54 & 0.14|0.54 & 0.12|0.52 & \textbf{0.17}|\textbf{0.56} & & 0.10|0.47 & 0.15|0.54 & 0.14|0.52 & 0.16|0.55 & 0.13|0.52 & \textbf{0.23}|\textbf{0.59} \\ 
RU$\rightarrow$EN & 0.31|0.62 & 0.32|0.63 & 0.32|0.63 & 0.32|0.63 & 0.33|0.64 & \textbf{0.35}|\textbf{0.65} & & 0.34|0.63 & \textbf{0.43}|\textbf{0.71} & 0.34|0.65 & 0.42|0.70 & \textbf{0.43}|\textbf{0.71} & 0.42|\textbf{0.71} \\
RU$\rightarrow$KK & 0.08|0.49 & 0.10|0.52 & \textbf{0.13}|0.53 & \textbf{0.13}|\textbf{0.54} & 0.12|\textbf{0.54} & \textbf{0.13}|\textbf{0.54} & & 0.15|0.55 & 0.21|0.61 & 0.18|0.58 & 0.22|\textbf{0.62} & 0.23|\textbf{0.62} & \textbf{0.24}|\textbf{0.62} \\ 
RU$\rightarrow$TR & 0.10|0.49 & 0.12|0.52 & 0.14|0.54 & 0.14|0.54 & 0.13|0.54 & \textbf{0.17}|\textbf{0.56} & & 0.11|0.49 & 0.16|0.56 & 0.16|0.55 & 0.18|0.57 & 0.16|0.55 & \textbf{0.22}|\textbf{0.60} \\
TR$\rightarrow$EN & 0.34|0.64 & 0.35|0.65 & 0.36|0.66 & 0.36|0.66 & 0.38|\textbf{0.67} & \textbf{0.39}|\textbf{0.67} & & 0.31|0.61 & \textbf{0.38}|\textbf{0.67} & 0.32|0.63 & \textbf{0.38}|0.66 & 0.36|0.66 & 0.37|0.66 \\
TR$\rightarrow$KK & 0.07|0.45 & 0.10|0.51 & \textbf{0.13}|\textbf{0.54} & \textbf{0.13}|\textbf{0.54} & 0.12|0.53 & \textbf{0.13}|\textbf{0.54} & & 0.08|0.46 & 0.14|0.53 & 0.14|0.52 & 0.16|0.55 & 0.14|0.53 & \textbf{0.19}|\textbf{0.57} \\
TR$\rightarrow$RU & 0.15|0.48 & 0.17|0.51 & 0.18|0.52 & 0.19|0.53 & 0.20|\textbf{0.54} & \textbf{0.21}|\textbf{0.54} & & 0.17|0.50 & 0.23|0.56 & 0.20|0.54 & 0.24|0.57 & 0.23|0.57 & \textbf{0.26}|\textbf{0.58} \\
\midrule
\textbf{Average} & 0.18|0.53 & 0.20|0.56 & 0.22|0.58 & 0.22|0.58 & 0.23|0.58 & \textbf{0.25}|\textbf{0.59} & & 0.20|0.55 & 0.27|0.61 & 0.23|0.59 & 0.27|0.62 & 0.26|0.61 & \textbf{0.30}|\textbf{0.63} \\
\bottomrule
\end{tabularx}
\caption{\ac{bleu}|\ac{chrf} scores for models on the FLoRes and KazParC test sets}
\label{tab:bleu_results}
\end{table*}

Conversely, in the ``KK$\rightarrow$'' translation direction, Google retained its translation accuracy predominance across both test sets, albeit with occasional instances where \texttt{parc} and \texttt{parsync} surpassed Google's performance. Notably, both \texttt{parc} and \texttt{parsync} consistently demonstrated the second-best performance, often matching or surpassing that of Yandex in this specific translation direction.

Within translation pairs involving Russian as the source language, out of the two models trained on our parallel corpus, \texttt{parsync} exhibited a consistent presence among the top three performers. Google, on the other hand, occasionally ceded its position to \texttt{parc} and Yandex in the RU$\rightarrow$EN language pair.

For the ``TR$\rightarrow$'' translation direction, \texttt{parsync} achieved noteworthy success, securing a leading \ac{bleu} score of 0.38 on the KazParC test set for TR$\rightarrow$EN and a commanding \ac{bleu} score of 0.13 in the TR$\rightarrow$KK language pair on the FLoRes test set, with Google being the frontrunner.

After thoroughly assessing the qualitative and quantitative results, we determined that the \texttt{parsync} model, fine-tuned on a combination of the KazParC corpus and synthetic data, displayed the highest results among the three developed models. In the upcoming Discussion section, we will simply refer to this model as ``Tilmash'' [t\textsci l\textprimstress m\textscripta \textesh], a Kazakh term denoting ``interpreter'', ``translator''.

It is worth noting that when comparing the translation results between \texttt{base} and Tilmash on the control language pairs, the latter displayed less favourable results, hinting at a possible decline in translation quality after fine-tuning (see Table~\ref{tab:control_pair_results}).

\begin{table}[!h]
% \fontsize{9}{10.8}\selectfont
% \setlength\tabcolsep{0.24cm}
\begin{tabularx}\columnwidth{cccccc}
\toprule
\multirow{2}{*}{\textbf{Pair}} &  \multicolumn{2}{c}{\textbf{\ac{bleu}}} & & \multicolumn{2}{c}{\textbf{\ac{chrf}}} \\
\cmidrule{2-3} \cmidrule{5-6}
& \texttt{base} & Tilmash & & \texttt{base} & Tilmash \\
\midrule
DE$\rightarrow$FR & 0.33 & 0.28 &  & 0.61 & 0.58 \\
FR$\rightarrow$DE & 0.22 & 0.19 &  & 0.55 & 0.53 \\
DE$\rightarrow$UK & 0.15 & 0.04 &  & 0.49 & 0.36  \\
UK$\rightarrow$DE & 0.19 & 0.16 & & 0.53 & 0.50 \\
FR$\rightarrow$UZ & 0.06 & 0.02 &  & 0.48 & 0.31 \\
UZ$\rightarrow$FR & 0.25 & 0.22 &  & 0.56 & 0.53 \\
\bottomrule
\end{tabularx}
\caption{Results of the \texttt{base} and Tilmash models on the control language pairs on the FLoRes test set}
\label{tab:control_pair_results}
\end{table}

\begin{table*}[!h]
\setlength\tabcolsep{0.4cm}
\begin{tabularx}\textwidth{cclcc}
\toprule
\textbf{Pair} & \textbf{Type} & \multicolumn{1}{c}{\textbf{Text}} & \textbf{\ac{bleu}} & \textbf{\ac{chrf}} \\
\midrule
\multirow{4}{*}{KK$\rightarrow$EN} & source & \textit{\scelll{Ыстық және желді.\\Ystyq jane jeldi.}} & & \\
& reference & \textit{It is hot and windy.} & 1.00 & 1.00 \\
\cdashline{2-5}
& Tilmash & \textit{It's hot and windy.} & 0.55 & 0.81 \\
& Yandex & \textit{Hot and windy.} & 0.00 & 0.66 \\
& Google & \textit{Hot and windy.} & 0.00 & 0.66 \\
\hline
\multirow{4}{*}{KK$\rightarrow$EN} & source & \textit{\scelll{1 қыркүйекте бесінші ана өлімі тіркелді.\\1 qyrkuiekte besinshi ana olimi tirkeldi.}} & & \\
& reference & \textit{On September 1, the fifth maternal death was registered.} & 1.00 & 1.00 \\
\cdashline{2-5}
& Tilmash & \textit{A fifth maternal death was recorded on 1 September.} & 0.27 & 0.63 \\
& Yandex & \textit{On September 1, the fifth maternal death was registered.} & 1.00 & 1.00 \\
& Google & \textit{On September 1, the fifth maternal death was recorded.} & 0.81 & 0.86 \\
\bottomrule
\end{tabularx}
\caption{A selection of translation outputs from Tilmash, Yandex, and Google}
\label{tab:result_comparison}
\end{table*}

The lower \ac{bleu} scores for Kazakh and Turkish translations can be attributed to the agglutinative nature of these languages. In agglutinative languages, words are formed by stringing together different morphemes, leading to longer and more complex words. This linguistic characteristic poses a challenge for translation models, as they may have difficulty capturing the complicated morphological structures, resulting in a statistically lower \ac{bleu} score.

However, we observed that the \ac{chrf} score remains relatively stable across language pairs. This suggests that the overall translation quality, measured by \ac{chrf}, is consistent across all language pairs. The \ac{chrf} metric considers n-grams at the character level and provides a more robust evaluation that is less sensitive to the structural differences between languages.

We hypothesise that the differences in translation quality between language pairs may be influenced by the resourcefulness of the languages and the training data available for the baseline \ac{nllb} model. Languages with richer linguistic resources and diverse training data may demonstrate better translation results.

\section{Discussion}\label{sec5}
The comparison of the results of Tilmash with those of Yandex and Google on the FLoRes and KazParC test sets reveals that the performance of our model is on par with that of the industry giants.
It is particularly pleasing to note that Tilmash yields consistent results on the diverse FLoRes test set, spanning a wide range of topics, from rare diseases to long-extinct dinosaurs, which may not be present in KazParC. This further reinforces the versatility of our model in effectively translating texts across various domains. That said, Tilmash appears to struggle with translating figurative expressions, such as proverbs and idioms, where conveying both literal accuracy and the rich cultural, historical, and emotional connotations they hold can be a challenging balance to maintain.

While it is true that the results of Tilmash are not significantly higher than those of \texttt{parc}, which was exclusively trained on our parallel corpus and, in some cases, even lower (see, for instance, ``$\rightarrow$EN''), we must acknowledge that the inclusion of synthetic data in the training set has had a positive impact on the performance of Tilmash, as evident from its strong performance on the FLoRes test set—a feat that the \texttt{parc} model cannot claim.
The substantial increase in the number of word types, and, consequently, the diversity of vocabulary, introduced by the synthetic data not only appears to enhance translation performance but also suggests the potential of utilising synthetic data in conjunction with much smaller amounts of human-translated parallel data to achieve improved results. However, it is important to remain mindful of the inherent translation inaccuracies and incorrect syntactic structures that can result from \ac{mt} of large, web-crawled, and uncurated data.
For example, Tilmash occasionally stumbles over second-person singular pronouns in Kazakh (\textit{сіз}, \textit{сен}), Russian (\textit{вы}, \textit{ты}), and Turkish (\textit{siz}, \textit{sen}) when translating the English ``you''. This can lead to instances where Tilmash produces informal (\textit{сен}, \textit{ты}, \textit{sen}) pronouns instead of the expected polite (\textit{сіз}, \textit{вы}, \textit{siz}) forms. We attribute this issue to the use of the synthetic corpus, as \texttt{parc}, trained solely on KazParC, accurately handles these pronouns.

A thorough examination of the performance of Tilmash, Yandex, and Google across the domains within the KazParC test set reveals the remarkable superiority of Tilmash in legal documents and texts pertaining to the general domain.\footnote{Due to space constraints, we have published the detailed tables of results per domain on our GitHub page.} This notable performance is observed in nine translation directions, as indicated by either \ac{bleu} or \ac{chrf} scores, which we attribute to the extensive presence of well-translated legal documents and everyday social expressions within the parallel corpus (see Table~\ref{tab:kazparc_domain_stats}). 
The somewhat lower, yet still comparable, results observed in the mass media domain, despite the majority of texts in KazParC originating from this domain, can be attributed to several factors. It is challenging to rival Google and Yandex in this domain, as their models are likely to have been extensively trained on news articles. Additionally, the presence of numerous proper nouns (e.g., names of individuals, organisations, locations, and more) and abbreviations within news content can pose challenges for \ac{mt} models in ensuring accurate handling.

Table~\ref{tab:result_comparison} provides some examples of KK$\rightarrow$EN translation. We can see that in the first example Tilmash demonstrated a distinct approach compared to Yandex and Google, which simply translated the adjectives into English. Not only was Tilmash able to correctly detect that the source sentence was an impersonal construction, but it also produced ``it'', which effectively functions as a placeholder for the weather condition. While the \ac{bleu} and \ac{chrf} scores are not perfect, it is worth emphasising that the difference between the reference sentence and the Tilmash-generated sentence solely lies in the use of the contraction ``it's'', with both sentences conveying the same information and maintaining identical grammatical structures.

In the second example, we observe that the sentence generated by Tilmash, as well as the reference sentence and those produced by Yandex and Google, convey similar meanings but exhibit differences in sentence structure, word choice influenced by regional date conventions (``September 1'' vs. ``1 September'') and formality (``registered'' vs. "recorded"), and the use of articles (``the'' vs. ``a''). While, in many contexts, these variations in dates and verbs can be used interchangeably, the choice of articles depends on contextual information. Specifically, it hinges on whether one is referring to one of multiple maternal deaths or a specific, previously mentioned, or contextually precise fifth maternal death. Without context, Tilmash may face challenges in determining the appropriate article to use while maintaining proper grammar. Nevertheless, we believe that such cases can be effectively addressed by a human translator during the post-editing phase, if necessary.

\vspace{-0.1cm}

\section{Conclusion}\label{sec6}

We have introduced KazParC, a parallel corpus developed for \ac{mt} of Kazakh, English, Russian and Turkish. It is the first and largest publicly available corpus of its kind and includes 371,902 parallel sentences from different domains created with the help of human translators. In addition, our research has led to the development of the Tilmash \ac{nmt} model, which has demonstrated remarkable performance, often matching or surpassing Yandex Translate and Google Translate, as evidenced by standard evaluation metrics such as \ac{bleu} and \ac{chrf}. Both KazParC and Tilmash are available for download under the Creative Commons Attribution 4.0 International Licence (CC BY 4.0) from our GitHub repository.\textsuperscript{\ref{ft:github}}

In the future, we are committed to expanding KazParC to cover a wider range of domains and lexica, including figurative expressions, with the aim of improving translation quality. We also plan to conduct further experiments with the \ac{nllb} model to preserve the original translation quality in non-target language pairs. In addition, we will continue to explore different pre-trained models and training parameters to refine our models.

\section{Acknowledgements}\label{sec:acknowledgements}
We extend our sincere gratitude to Aigerim Baidauletova, Ainagul Akmuldina, Assel Mukhanova, Aizhan Seipanova, Askhat Kenzhegulov, Elmira Nikiforova, Gaukhar Rayanova, Gulim Kabidolda, Gulzhanat Abduldinova, Indira Yerkimbekova, Moldir Orazalinova, Saltanat Kemaliyeva, and Venera Spanbayeva for their invaluable assistance with translation throughout the entirety of the study.

\section{Bibliographical References}\label{sec:references}
\bibliographystyle{references_style}
\sloppy
\bibliography{references}

\end{document}